%% file: root.tex
\documentclass[letterpaper, 10 pt, conference]{ieeeconf}  
\usepackage{tikz}
\usepackage{hyperref}

\usepackage{amsmath,amssymb,mathrsfs}

\IEEEoverridecommandlockouts                              

\title{\LARGE \bf
PANDORA: The Open-Source, Structurally Elastic Humanoid Robot
}

\author{
    Connor W. Herron $^1$,\and Alexander J. Fuge $^1$,\and Benjamin C. Beiter $^1$,\and Zachary J. Fuge $^1$,\and Nicholas J. Tremaroli $^1$,\and Stephen Welch $^1$,\and Maxwell Stelmack $^1$,\and Madeline Kogelis $^1$,\and Philip Hancock $^1$,\and Ivan Fischman Ekman Simoes $^1$,\and Christian Runyon $^1$, Isaac Pressgrove $^1$, Alexander Leonessa $^1$
\thanks{*This work was not supported by any organization}
\thanks{$^{1}$ All Authors are with the Department of Mechanical Engineering, Virginia Polytechnic Institute and State University in Blacksburg VA, USA.}%
\thanks{$^{2}$ Corresponding Author: Connor Herron, {cwh@vt.edu}}}%

\begin{document}

\maketitle
\thispagestyle{plain}
\pagestyle{plain}

\begin{abstract}

In this work, the novel, open-source humanoid robot, PANDORA, is presented where a majority of the structural elements are manufactured using 3D-printed compliant materials. As opposed to contemporary approaches that incorporate the elastic element into the actuator mechanisms, PANDORA is designed to be compliant under load, or in other words, structurally elastic. This design approach lowers manufacturing cost and time, design complexity, and assembly time while introducing controls challenges in state estimation, joint and whole-body control. This work features an in-depth description on the mechanical and electrical subsystems including details regarding additive manufacturing benefits and drawbacks, usage and placement of sensors, and networking between devices. In addition, the design of structural elastic components and their effects on overall performance from an estimation and control perspective are discussed. Finally, results are presented which demonstrate the robot completing a robust balancing objective in the presence of disturbances and stepping behaviors.

\end{abstract}


\input{Sections/Section1_Introduction}

\input{Sections/Section2_WholeBodyDesign}

\input{Sections/Section3_StructuralElasticity}

\input{Sections/Section4_ImpactOnControls}

\input{Sections/Section5_WalkingResults}

\input{Sections/Section6_Conclusion}
\vspace{-0.2cm}

\bibliographystyle{unsrt}
\bibliography{asmeconf-bib} 

\end{document}

%% file: Sections/Section1_Introduction.tex
\section{Introduction}
Humanoid robots are an increasingly prevalent technology with entertainment, industry, healthcare, and personal use applications \cite{carros2020exploring, jeong2022robot, dafarra2024icub3}. This initiative has been enhanced by the numerous humanoid robots founded by well-known groups and companies such as Agility, Tesla, Apptronik, Boston Dynamics, Boardwalk Robotics, and Figure AI \cite{ieee2023humanoids}. While these groups differ in use case, there is a general consensus that these robots will complete dangerous or monotonous, repetitive tasks that otherwise, a human worker would need to complete. Academically, humanoid robots, and specifically the design and control of under-actuated, bipedal robots, has been a common topic of research of several decades \cite{knabe2015design,  kajita2003resolved}. 

\begin{figure}[t]
\centering
\centering
\includegraphics[width=\linewidth]{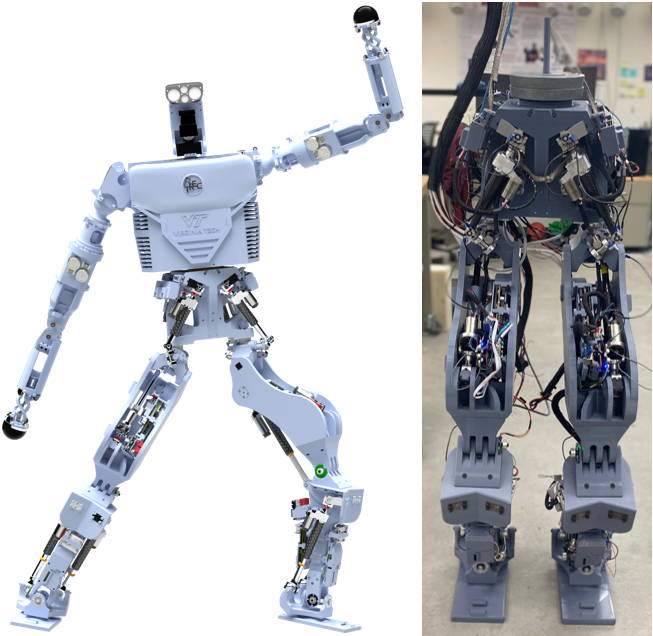}
\caption{PANDORA Humanoid Robot. Available at \href{https://gitlab.com/trec-lab}{gitlab.com/trec-lab}}	
\label{fig:main}
\vspace{-0.5cm}
\end{figure}

One consistent barrier to such research, however, is the cost and investment required to build a humanoid robot. For full-sized humanoid robots, significant funding is required to manufacture highly-complex components at precision machining tolerances along with the numerous sensors and computer components. Towards lowering the barrier to entry, there are a few open source options for humanoid and bipedal research platforms \cite{metta2008icub, ramos2021hoppy}. In this paper, we present the full-body design of our open-source humanoid robot PANDORA, shown in Figure \ref{fig:main}, whose structural components exhibit an elasticity from the compliant 3D printing materials. Elastic elements in the robot's actuation mechanism has been found to protect against shock impulses, increase efficiency, and increase control bandwidth \cite{pratt1995series, hopkins2015embedded}. In the author's experience, a dedicated compliant mechanism is unnecessary when utilizing compliant materials available in additive manufacturing methods. However, this benefit comes at the expense of increasing controls and estimation complexity which is further discussed in this work.



\begin{figure}[!t]
\centering
\includegraphics[width=\linewidth]{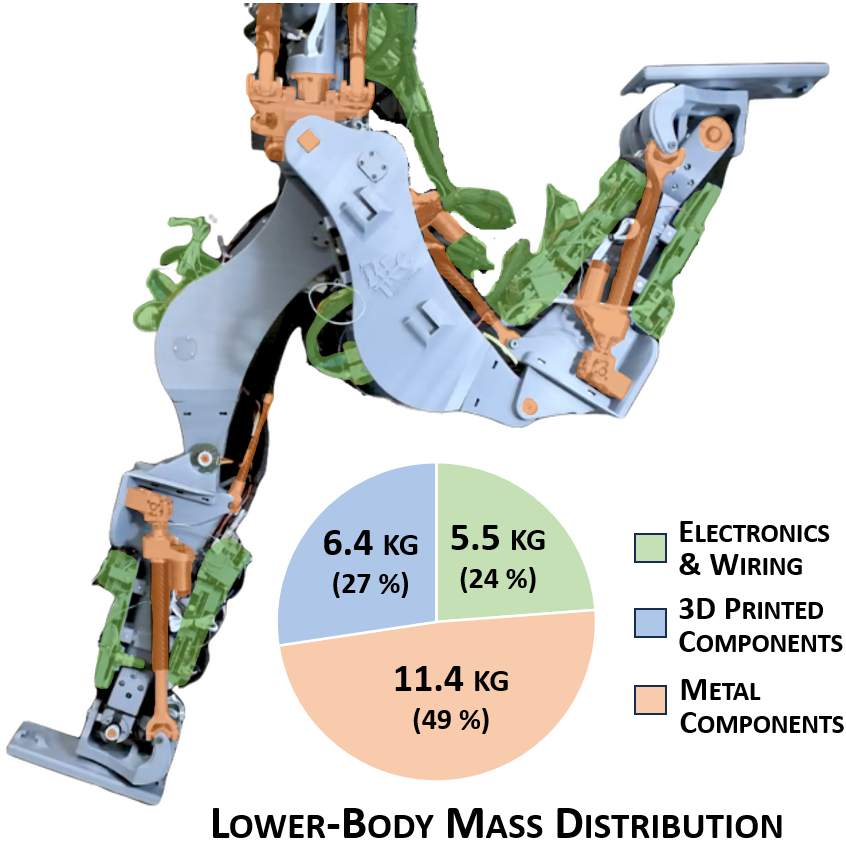}
\caption{The lower-body contains a combination of metal and 3D-printed plastic components.}
\label{fig:mass-distribution}
\vspace{-0.5cm}
\end{figure}

The contributions of this work include an overview of the humanoid robot PANDORA, an analysis of the compliant 3D printed components, a discussion on how structural elasticity affects standard humanoid robot control, and results of PANDORA completing balance behaviors in the presence of disturbances. To help make humanoid robotics research more accessible, PANDORA's entire design including the mechanical design CAD step-files, low-level PCB designs and firmware, and high-level software is provided at the TREC lab's open-source GitLab page, \href{https://gitlab.com/trec-lab}{gitlab.com/trec-lab}.

The rest of the paper will be structured as follows: Section 2 will present the full design of the robot; Section 3 will detail the structural elasticity resulting from the compliant components; Section 4 will discuss the impact that structural elasticity has on controller performance; Section 5 presents results of the robot balancing in the presence of disturbance and stepping behaviors; and Section 6 concludes the paper. 


%% file: Sections/Section2_WholeBodyDesign.tex
\section{PANDORA Design}
The PANDORA humanoid robot utilizes a hybrid approach of additive manufacturing (AM) and subtractive manufacturing (SM) methods to create a structure which is easy to fabricate and assemble. In addition, AM methods can easily manufacture compliant structural components which allows us to remove the elastic component typically designed internal to the actuator mechanism.
The joints are driven via linear actuators which rely on several methods of sensor feedback that are utilized for measuring the state of the robot. The sensor measurements and actuator motor commands are handled by a low-level controller which is a hardware component capable of managing two sets of actuator and joint pairs. The low-level controllers also contain the force control algorithm for the actuators where the desired force commands come from the high-level controller. Using EtherCAT communication, the state measurements are collected into the high-level controller contained in the TREC Robotics Software (TRS) which interfaces directly with the IHMC Open Robotics Software (ORS) platform. The ORS stack contains the planner and Whole-Body Controller (WBC) outputting desired impedance trajectories for the actuators to track. 

\begin{figure}[!t]
\centering
\includegraphics[width=\linewidth]{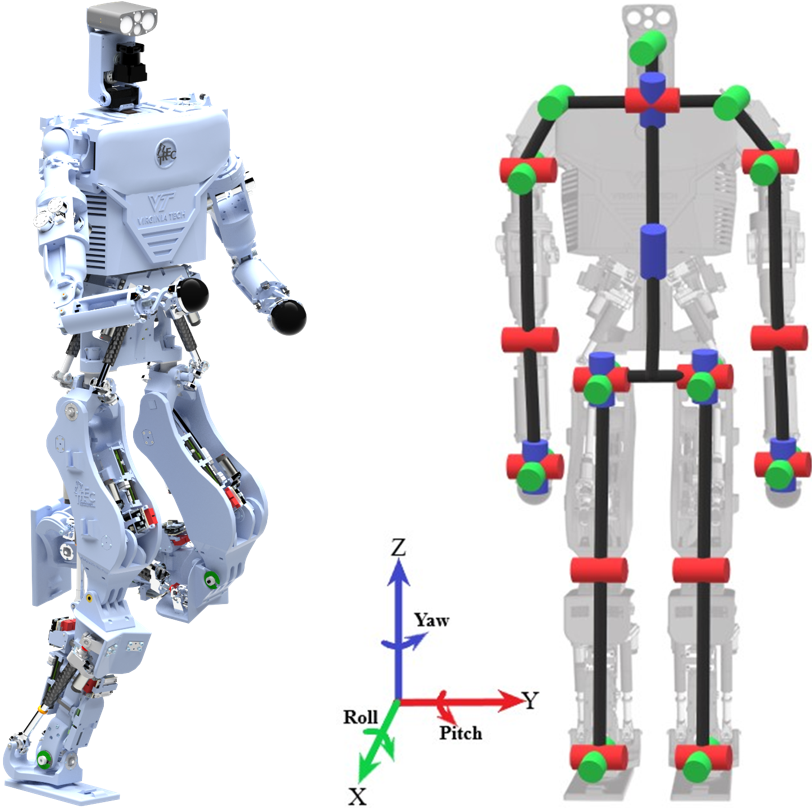}
\caption{PANDORA's kinematic design including range of motion capabilities allow for dynamic behaviors.}
\label{fig:mechanical_design}
\vspace{-0.5cm}
\end{figure}


\subsection{Mechanical Design}
The mechanical design of PANDORA leverages AM for most of the structural components which allows for a reduction in overall weight and assembly time. Traditional SM methods are unable to manufacture complex geometries resulting in robot designs with numerous components fastened together with bolts. However, AM allows multiple components to be combined, which significantly lowers cost, build and assembly time, and simplifies the overall design.

PANDORA stands at 1.9 meters and weighs 49 kg, which benefits from AM's ability to reduce part count and create complex geometries. Previous generations of humanoid robots at TREC, THOR and ESCHER \cite{knabe2015design}, were created using SM methods and had lower-body part counts of 510 and 480, respectively. In contrast, PANDORA’s use of AM has reduced the part count to just 228 in the lower body, achieving over a 50\% reduction from predecessors.
PANDORA's hybrid mechanical design employs AM for a majority of the structural components with a few SM parts in the critical joints and actuators. The lower body mass distribution is displayed in Fig. \ref{fig:mass-distribution}, where the AM parts account for 6.4 kg out of 23.27 kg or 27.5 \%. This design choice streamlined the assembly process, reducing the assembly time for the lower body to just 8 hours with two people.

Designed as a general-purpose robot, PANDORA requires a range of motion comparable to that of a human. It features 12 degrees of freedom (DoFs) in the lower body, 4 in the chest and head, and 14 in the arms, mirroring the capabilities of an average human. This is illustrated in Fig. \ref{fig:mechanical_design} where each of the roll, pitch, and yaw joints are highlighted.

The main structural components of the lower body are the pelvis, thigh, and shin, all manufactured using a Creality CR-10 3D Printer. These components are designed to maximize strength and minimize weight, with careful attention to load distribution and force management from the actuators. The pelvis supports five actuators for controlling 5 DoFs, houses various electronics, and handles weight distribution from the upper to the lower body. The thigh's curved structure enhances the range of motion for the hip and knee joints housing 2 actuators. Utilizing the ability of AM, there is an incorporation of slide rails for electronic mounting and calibration slots for encoder zeroing. The shin accommodates actuators for the ankle joint and various sensors, with a central beam structure to manage weight and forces during movement. 
\begin{figure*}[!t]
\centering
\includegraphics[width=\textwidth]{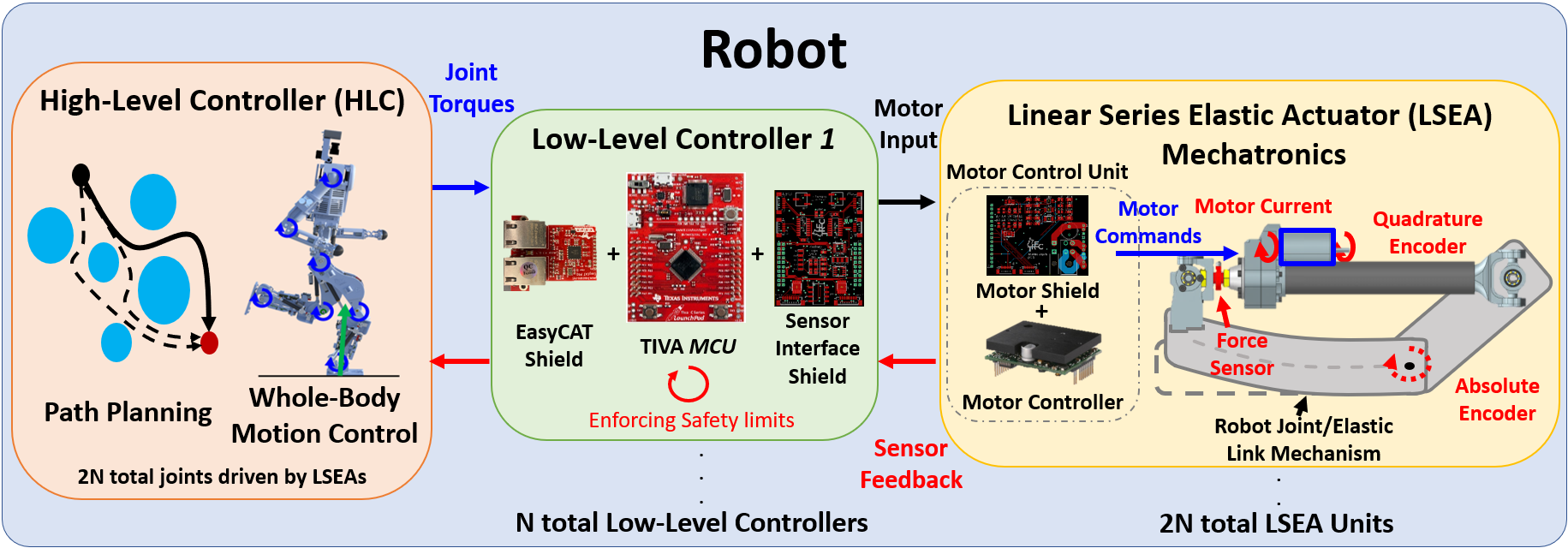}
\caption{Complete Overview for Robot Control Architecture.}
\label{fig:control-overview}
\vspace{-0.5cm}
\end{figure*}

The structural components of PANDORA's upper body consist of four main groups: the chest, the head and neck, and the left and right arms. The chest features a shelving system designed to hold two batteries, a mini PC, a network switch, and the shoulders. The chest structure is primarily composed of large 3D-printed plates, which are stabilized and supported by aluminum rails, ensuring a robust yet lightweight frame. In addition, the chest contains a mounting mechanism at the bottom to allow for yaw rotation of the upper body with respect to the pelvis.

The head and neck are mounted directly to the chest using two direct drive motors, providing stability and precise control. The head includes a Hokuyo LiDAR capable of rotation and a Carnegie Robotics Multisense S7 sensor, both of which are mounted on a 3D-printed head structure. This setup allows for comprehensive scanning of the surrounding environment for perception purposes.

The arm design takes a bio-engineered approach, inspired by human anatomy. The shoulder houses two motors, with a third motor mounted in the chest and connected via a pulley system for enhanced control. The humerus is designed to mount four motors, which control the elbow and the 3 DoFs found in the wrist. The elbow is engineered such that the humerus and forearm meet in a double joint configuration, enabling the forearm to rotate around the humerus. The forearm itself is designed to house all the necessary cabling for the 3 DoFs and mechanisms that provide the wrist with its yaw rotation. The wrist structure, composed of intricately designed AM parts, mimics the movements of a human wrist, offering a high degree of dexterity and flexibility.

This innovative structural design leverages the strengths of both additive and subtractive manufacturing. AM allows for the creation of complex geometries and significant weight reduction, while SM components provide the necessary strength and precision for smooth, reliable operation. This hybrid approach ensures that PANDORA remains a versatile and robust platform, capable of performing a wide range of tasks with human-like agility and responsiveness. More details regarding PANDORA's lower body can be found in \cite{fuge2023design} where two additional publications are being prepared on the upper body design and the best practices for 3D printing large-scale robotic systems.

\subsection{Joint and Actuator Design}

The joints of PANDORA utilize a hybrid design that combines SM and AM methods. This approach ensures the smooth transfer of forces and rotational movements with lower backlash at the joints (compared to a full AM approach), while also withstanding the significant stresses experienced at the joints (areas where SM components are particularly advantageous). The hip roll/pitch and knee pitch joints exhibit some backlash which, if too much, can lead to instability in the joint or whole-body controllers.

In PANDORA's lower body, the joint configuration includes 2 DoFs in the ankle, 1 DoF in the knee, and 3 DoFs in the hip, with an additional DoF in the pelvis for upper body rotation. All joints in the lower body are driven by repurposed linear actuators from THOR and ESCHER \cite{knabe2015design}. The joint designs utilize a mix of SM and AM parts, with common components being bearings pressed into aluminum holders that screw into the plastic. These aluminum holders are used 16 times across the 4 DoFs in the hip and ankle. The knee employs bushings to facilitate smooth rotation. Specially designed aluminum gimbals in the hip and ankle help transfer weight and forces smoothly between the joints. This design, aided by linear actuators, allows PANDORA to achieve a range of motion similar to that of the human body.

The upper body features a combination of direct-drive and cable-driven motors. The cable-driven design ensures that more weight is kept close to the body rather than at the extremities. Unlike the lower body, which primarily uses SM joints, the main connections in the arms are all AM to minimize weight. The wrist is assembled using AM-created pins that snap into place with unique curved connection bars that allow for 2 DoFs. Moving up, a double-jointed elbow maintains a constant distance throughout its motion, keeping the controlling cable for the wrist taut. The shoulder incorporates a direct-drive motor for yaw rotation and a 3D-printed ball joint for 2 DoFs, controlled by cabling. Additionally, the upper shoulder includes a roll joint with another integrated into the chest for an enhanced range of motion. This design was inspired by the human Serratus Anterior muscle, which assists in shoulder movement. The arm's design aims to mimic human anatomy, providing PANDORA with a reach similar to that of a human.

The hybrid use of SM and AM in the joint design ensures that PANDORA maintains the structural integrity required for smooth, precise movements while leveraging the weight-saving benefits of 3D printing. This innovative approach allows for a robust and flexible humanoid robot capable of performing a wide range of tasks.

\subsection{Electrical and Networking}
As displayed in Fig. \ref{fig:control-overview}, an electrical system has been designed entirely in-house with a focus on modularity. The lower body is controlled using Low-Level Controllers (LLC), which is a hardware component composed of an EasyCat shield, a TIVA microcontroller, and a Sensor Interface shield. Each LLC can communicate and control two actuators to drive two series or parallel joints. The LLC is responsible for collecting joint (absolute) and motor (quadrature) encoder, motor current, and force sensor feedback at 1000 Hz. A variety of hardware filters are utilized for conditioning the motor current and force sensor feedback to remove the 60 Hz mains hum and high-frequency motor induced signal noise. At the left and right ankle LLC's, the ground reaction forces and torques are measured via ATI FT sensors. In addition to sensor feedback collection, the LLC sends PWM commands to the Motor Control Unit (MCU) for controlling an actuator. Each LLC contains a force impendance controller based on a disturbance observer approach for tracking the desired torques which output from the higher-level controller. Details regarding this joint level control strategy will be the focus of a future publication. 

In order to control PANDORA for full-body behaviors, the sensor feedback is communicated to the high-level controller at a rate of 500 Hz. This rate can be increased by lowering the communication bandwidth, but this rate is sufficient for robust balancing and walking behaviors. Networking with the high-level controller is completed via EtherCAT using the EasyCAT Pro shield on the LLC. PANDORA's 6 lower body LLCs are arranged in a daisy-chained master-slave configuration where a central computer containing the high-level controller operates as the master. Data transfer sychronization is critically important for stable and robust control. Our prior work introduced custom-design communication protocols build on top of etherCAT and utilized a Master Process ID (MPID) for monitoring the synchronization of each embedded system \cite{tremaroli2023adaptive}. For safety purposes, each LLC checks a variety of sensor feedback conditions to determine safe operation (e.g. min/max joint encoder and force sensor) which can trigger a HALT state turning all the robot's motors off. In addition, the LLC firmware has been abstracted such that sensors can effortlessly be added or removed with a developer GUI \cite{tremaroli2023adaptive}.

Distributed or modular control systems such as these depend on accurate timing constraints for proper executution. The LLCs are responsible for collecting, conditioning, and communicating the sensor feedback to the high-level controller (central computer), checking safety conditions, and controlling the linear actuators to track the desired joint trajectories. Stability cannot be guaranteed if any of these processes are unable to be executed before a cycle period expires. As discussed in \cite{stelmack2024satisfying}, a real-time operating system (RTOS) is applied on top of the existing TIVA microcontroller firmware for enforcing and verifying timing constraints. More details regarding the electrical and networking approach are presented in \cite{herron2023design}.

\subsection{High-Level Control}
As displayed in Fig. \ref{fig:control-overview}, a high-level controller is responsible for utilizing sensor feedback from the robot to determine subsequent joint torques for executing desired whole-body behaviors. As shown in \cite{koolen2016design, hopkins2015compliant}, the high-level controller is composed of a series of planners including path, footstep, and CoM planners and a Whole-Body Controller (WBC). The Whole-Body Controller is an optimization problem designed to compute joint accelerations and torques based on desired motion tasks which correspond to a whole-body behavior. These tasks can include desired momentum rates of change, spatial accelerations, and joint accelerations which are orchestrated to represent complex whole-body behaviors such as balancing, manipulation, walking, and running \cite{koolen2016design, hopkins2015compliant}. On PANDORA, the high-level controller is designed in the TREC Robotics Software (TRS) \cite{trecTRS} which interfaces with IHMC's Open Robotics Software (ORS) \cite{ihmcORS}. TRS is framework with a variety of tools including communication, fault detection, controllers, and PANDORA's software starters. ORS contains the the footstep planner, DCM-based CoM planner, and WBC optimization problem which has been carefully hand-tuned for achieve robust control for PANDORA.


%% file: Sections/Section3_StructuralElasticity.tex
\section{Structural Elasticity}

\begin{figure}[t]
\centering
\includegraphics[width=\linewidth   ]{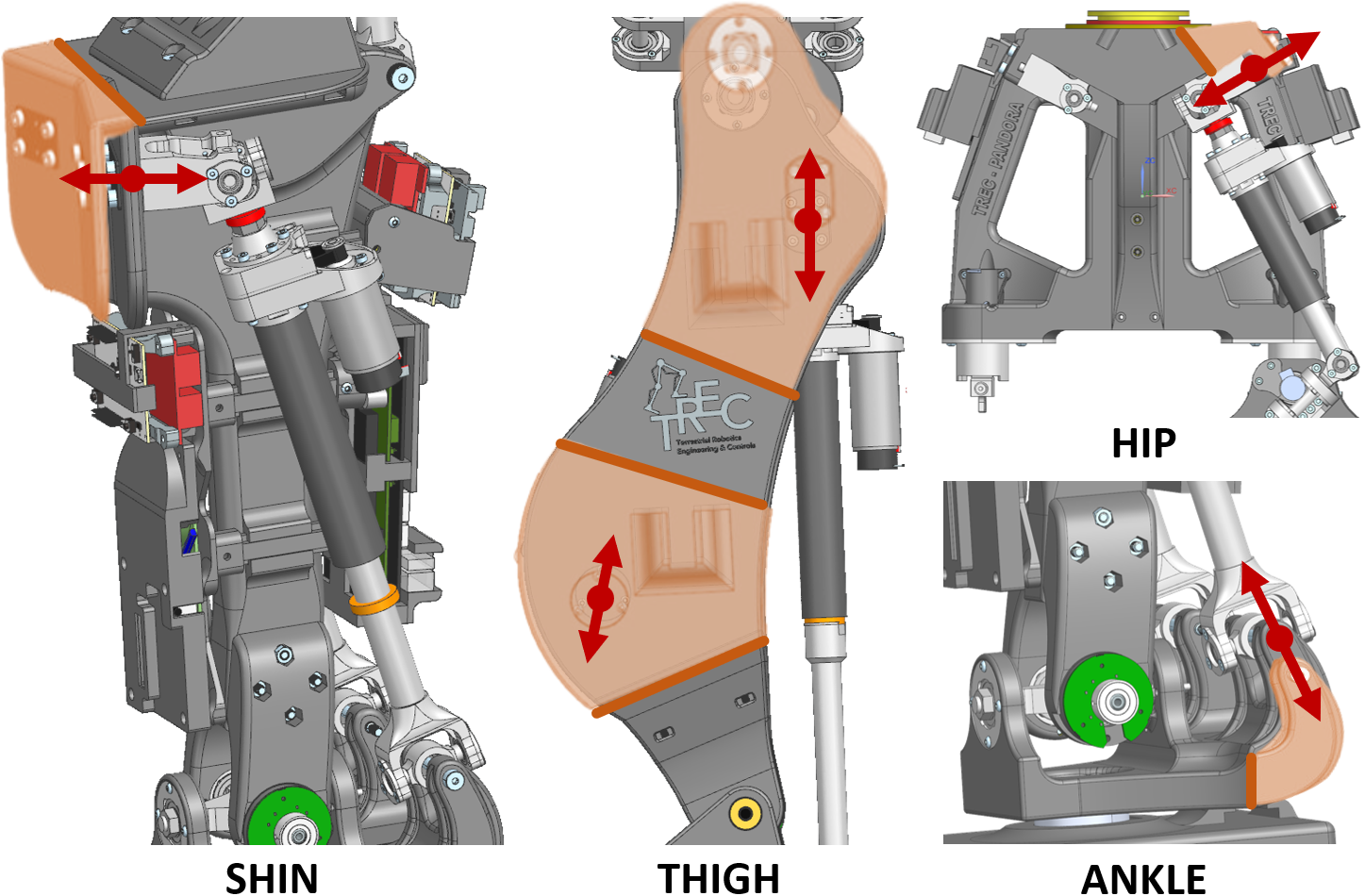}
\caption{Intended Structural elastic deflection (orange regions) are caused by the actuator loads (red arrows) and joint loads (not shown).}	
\vspace{-0.5cm}
\end{figure}

As discussed in \cite{pratt1995series, hopkins2015embedded}, elasticity offers shock protection to the mechanical actuation system, increased force/torque control bandwidth, and efficiency improvements from stored and returned energy. In the provided examples, a mechanical spring is featured as a component in series with the actuator which drives a joint of the robot. In this work, we use the term "structural elasticity" to capture the idea that the structural components are intentionally designed to be compliant. From a mechanical perspective, this approach reduces part count, design complexity, and cost \cite{fuge2023design}. From a controls perspective, however, too much compliance would lead to an infinite degree of freedom system because the linkages themselves are elastic making it extremely difficult or impossible to achieve stable behaviors. As discussed in our previous work \cite{fuge2023design}, the loading performance of 3D printed manufactured components can be affected by the design, material, speed, printer quality, and print settings. One of the most important design considerations is print orientation, where the best approach is to print perpendicular to the component's nominal load conditions. Thus, it can be quite difficult to simulate or understand the performance of a structural component prior to printing where a significant amount of research is still needed. Future work will provide details regarding this design and manufacturing process for achieving the best performance. In practice, we have found that structural elasticity significantly impacts joint-level control, state estimation, and whole-body control. 

To avoid complex modelling of the full-order system, the structural elasticity is compensated for at the joint-level controller. The joint-level controller tracks desired impedance (torque and position/velocity) set points from the high-level controller. Using a similar approach to \cite{hopkins2015embedded}, the actuator system utilizes a disturbance observer-based approach to track actuator force objectives using a simple inverse jacobian transpose of the desired torque objectives. The disturbance observer model is designed based on an ideal titanium leaf spring model which has been found empirically using an actuator test bed. For controlling PANDORA, however, the titanium leaf springs are removed relying on the structural components for compliance where the disturbance observer provides feedback to compensate for these modelling differences. This robust control approach also allows us to modify the structural design without modifying the joint-level controller, allowing for continuous iteration of the mechanical system. 

In addition to compensating the elasticity differences between the titatium leaf spring and the structural components, disturbance observers can remove stiction and help with stability in the presence of backlash. Backlash is one of the most prevalent and difficult components with 3D printing manufacturing, where it can be challenging to achieve concentric circles for bearings. In the author's experience, the disturbance observer has a tendency to excite the backlash which can lead to consistent chattering or even instability. Thus to avoid chattering due to backlash, a $0 \leq k_{dob} \leq 1$ gain is provided in the disturbance feedback loop. This gain is symmetrically applied across the robot where the hip and ankle actuators have a $k_{dob} = 0.8$, and the thigh/knee actuators have a $k_{dob} = 0.4$. This joint level control approach is capable of tracking desired joint torque and impedance setpoints at a real-time rate for balancing and walking objectives. Since the structural elasticity is unmodelled at the full-order model, however, this introduces challenges to state estimation and whole-body control performance.





%% file: Sections/Section4_ImpactOnControls.tex
\section{Impact on controls}




The process of migrating the joint level control strategy from an ideal pendulum test bed to the lower body multi-DOF system was challenging due to increased backlash and elasticity differences. Moderate tuning was applied to the PID controller and the impedance gains for each actuator carefully monitoring the trade-off between stability and high-speed control. An important part of joint control is the mapping of joint torques into actuator forces. As highlighted in \cite{KUMAR2020102367}, parallel-actuated robots typically are modelled as open chain robots where the output joint torque is simply cast to a linear force. In this case, we simply utilize an inverse jacobian mapping 
\begin{equation} \label{eqn:force-torque-mapping}
    \mathbf{f}_{\rm act} = \mathbf{J}^{\rm -T}(\mathbf{q}_{\rm joint})\boldsymbol{\tau}_{\rm joint}
\end{equation}
where $\mathbf{f}_{\rm act} \in \mathbb{R}^{2}$ is the actuator force, $\mathbf{q}_{\rm joint} \in \mathbb{R}^2$ are the pair of joints being driven, $\mathbf{J}(\mathbf{q}_{\rm joint}) \in \mathbb{R}^{2\times 2}$ is the mechanism jacobian, and $\boldsymbol{\tau}_{\rm joint} \in \mathbb{R}^2$ are the joint torques. As discussed in \cite{fuge2023design}, the actuators and joints are driven in pairs where the hip (roll/yaw) and ankle (roll/pitch) joints has both actuators contributing to both joints, and the hip and knee pitch joints are driven by single actuators. Therefore, the joint angle $\mathbf{q}_{\rm joint}$ is critical for proper mapping between the joint and actuator spaces to achieve stable joint control. Elastic bodies create a phase delay which, if unmodelled, can lead to instability. With respect to range of motion constraints, the actuator positions can uniquely be defined by a set of joint positions and vice-versa. For a static no-load condition, the motor encoder can provide the same kinematic mapping
\begin{equation} \label{eqn:actuator-position-estimates}
    \mathbf{q}_{\rm act} = f_{j,a}(\mathbf{q}_{\rm joint}) = f_{m,a}(\mathbf{q}_{\rm motor})
\end{equation}
where $\mathbf{q}_{\rm act} \in \mathbb{R}^{2}$ are the actuator positions, $ f_{j,a}(\mathbf{q}_{\rm joint})$ is the kinematic mapping from joint to actuator space, and $f_{m,a}(\mathbf{q}_{\rm motor})$ is the kinematic mapping from the motor to the actuator space. These position estimates can become out of phase resulting in desired force trajectories which cannot be tracked at the actuator controller level. Instead of simply utilizing an inverse jacobian for torque to force mapping, a flexible joint  model may improve stability \cite{albu2007unified}.

\begin{figure}[t]
\centering
\includegraphics[width=\linewidth]{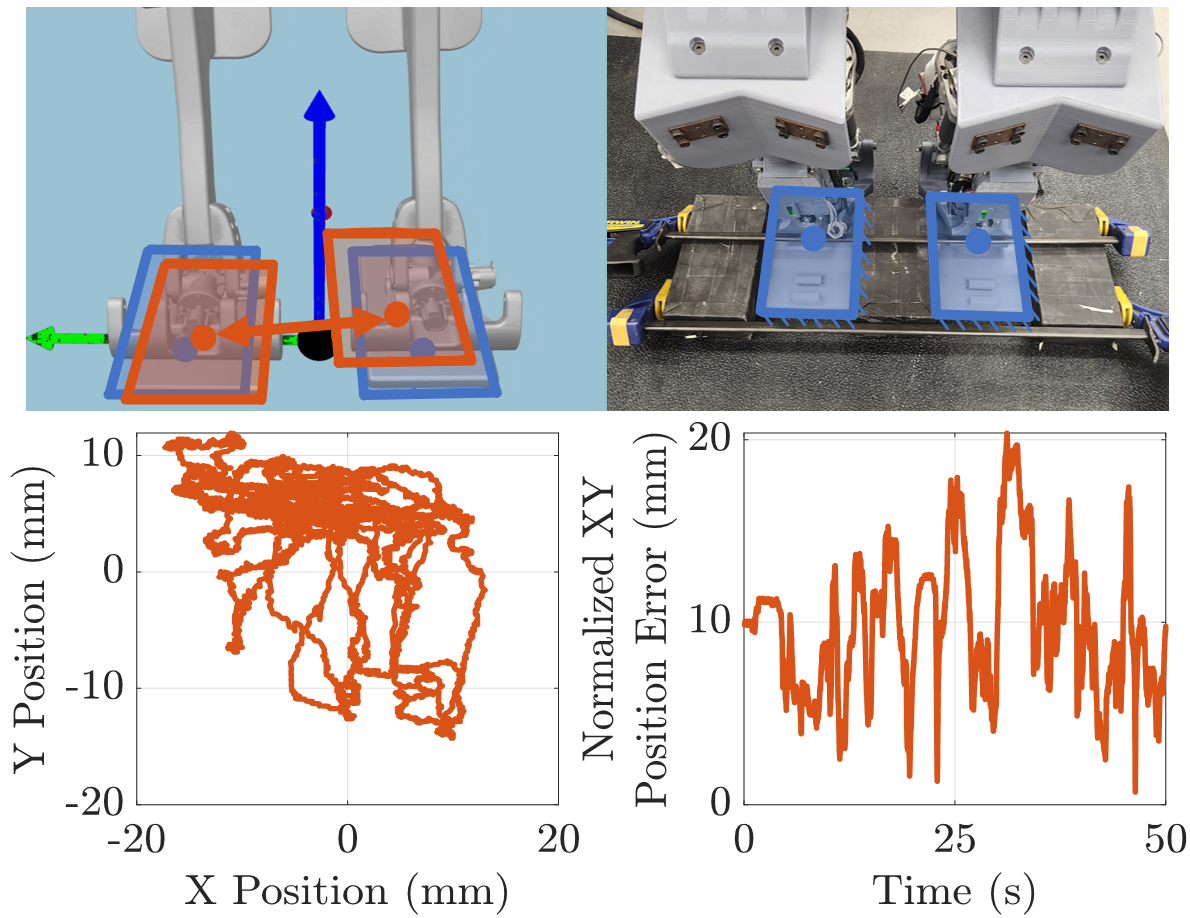}
\caption{Kinematic deflection on the right foot caused by applied motion during zero impedance}	
\label{fig:kinematic-deflection}
\vspace{-0.5cm}
\end{figure}

\subsection{State Estimation and WBC}
In addition to joint control concerns, State Estimation and WBC performance are critical for achieving stable whole-body behaviors such as robust balancing and walking. For legged robots, a critical component for defining their stability is based on the center of mass (CoM) position and velocity state estimate. This estimate is heavily dependent on the base orientation measurement and ground contact detection. 

As discussed in \cite{koolen2016design, hopkins2015compliant, englsberger2014overview, kajita2003resolved}, a planner determines CoM reference trajectories which are encoded into a centroidal momentum task for the WBC. Due to PANDORA's significant structural elasticity, the kinematic representation of rigid bodies does not always capture the robot's current state. An example of this kinematic error can be seen in Fig. \ref{fig:kinematic-deflection}, where the top right image displays that the real robot's feet have been fixed to the ground and the top left image displays the virtual robot's feet visualized from forward kinematics based on the sensor measurements. The bottom graph represents the XY distance between the right and left feet minus the known physical XY distance in the world frame. During the experiment, an operator is consistently applying forces to the pelvis to induce the forward kinematic error. 

During this motion, the right foot deflects by up to 2 cm where these differences can be visually discerned from the virtual robot. As discussed in Section \ref{sec:results}, the balancing test displays even worse kinematic error when dealing with disturbances and during contact changes when walking. Depending on the contact detection model, this can result in fast or instantaneous changes in CoM position which translates to significant momentum commands from the WBC which can easily destabilize the robot. This issue is discussed in  \cite{koolen2016design} (Section 6.6), introducing a small heuristic which adds a stiffness term to the joint position based on the applied torque. However, PANDORA's joint torques are not directly measured as in \cite{koolen2016design} and are estimated using the kinematic mapping based on the joint position measurements. As previously discussed, the difference between the motor and joint encoders can be significant. In addition, the force sensors values can be noisy and directly adding them to the joint position feedback does not always provide stable feedback. Thus, the authors are currently testing a Kalman filter-based approach which combines the motor and joint encoders, and force sensor to correct the joint position. This effort focuses on improving the kinematic accuracy and joint level stability margin.






%% file: Sections/Section5_WalkingResults.tex
\begin{figure*}[!t]
\centering
\includegraphics[width=\linewidth]{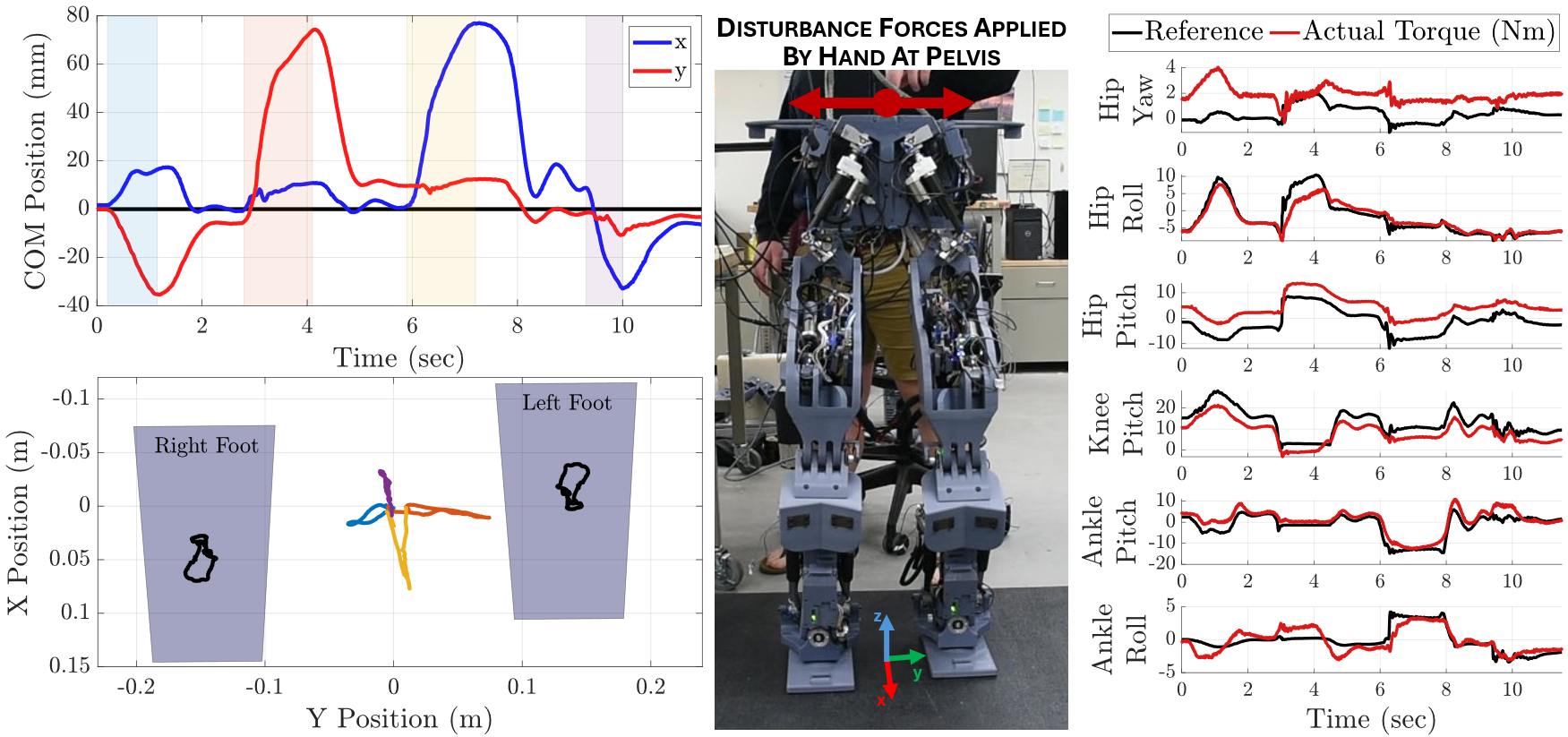}
\caption{Robust balance control of PANDORA under disturbances applied at the pelvis. Over the course of a 12 sec trial, forces were applied at the pelvis in the $-y$, $+y$, $+x$, and $-x$ directions. Left shows the center-of-mass displacement under the disturbances both over time and in the x-y plane. Black lines over the foot polygons indicate how much the feet were erroneously estimated to move over the course of the trial. Right shows joint torque tracking over the same period for the right leg. Reference torque is from the WBC and actual is the torque generated and measured by the linear actuators.}	
\label{fig:experimentImage}
\vspace{-0.5cm}
\end{figure*}

\section{Hardware Locomotion Results}



Despite the challenges to control presented by the structural elasticity, whole-body dynamic control of PANDORA can still be achieved for simple motion behaviors. This is demonstrated by two experiments. The first shows robust balance on two feet, even in the presence of disturbances at the pelvis. The second presents a stepping 'in-place' behavior, where the robot completes several steps in a row, even though the gait is not stable indefinitely. 

\subsection{Robust Balancing in the Presence of Disturbances} \label{sec:results}

Figure \ref{fig:experimentImage} shows the results of the double support balance task. In this experiment, PANDORA balances on two feet with the objective of keeping the CoM centered and stationary between the two feet. The robot is operating under WBC with 8 tasks: task-space position control of each foot (with a desired zero acceleration in stance), task-space orientation control of each foot, orientation control of the pelvis, whole-body linear momentum control, and a joint-space privileged position command biasing the knees to avoid singularity. This is combined with positive ground reaction force and dynamic feasibility constraints to generate desired joint accelerations and, through inverse dynamics, joint torques at every time-step. The controller is implemented in the open-source Stack-of-Tasks control software from IHMC's ORS \cite{ihmcORS}.

During the balance experiment, the disturbance forces are applied by hand to the top of the pelvis by an operator. As displayed in Fig. \ref{fig:experimentImage}, the disturbance force causes a displacement from zero of the CoM in the x-y plane. Once released, the balance control returns the CoM to an equilibrium. This recovery is true for a disturbance in any direction, including z, though that is not pictured. Fig. \ref{fig:experimentImage} also shows the variation of the estimated feet positions (black line), which varied up to 5 cm during the double stance balance experiment. While the whole-body and joint control approach is capable of maintaining stability to achieve the desired balance control behavior, this problem becomes more difficult during contact changes when stepping.

\subsection{In-Place Stepping}
The same controller is utilized in the second experiment, but now with the planned behavior of stepping in place. A state machine transfers the WBC tasks from double support to single support where the foot position tracking task commands a swing trajectory for the swing foot. A divergent component of motion (DCM) based LQR controller is used to drive the CoM to track a balanced trajectory along a predefined footstep plan. As the desired behavior is set to step in place, the desired footstep plan is always reset to match the current footstep positions.
When implementing this controller, it became clear that structural elasticity has a large impact on walking behaviors. This comes from a failure to track exact desired joint torque trajectories, errors in momentum estimation, and most impactfully, the kinematic differences in estimated feet position. When entering single stance during stepping, the stance leg is bearing the weight of the entire robot, increasing the errors in the state estimation, as previously displayed in Fig. \ref{fig:kinematic-deflection}. 

\begin{figure}[t]
\centering
\includegraphics[width=\linewidth]{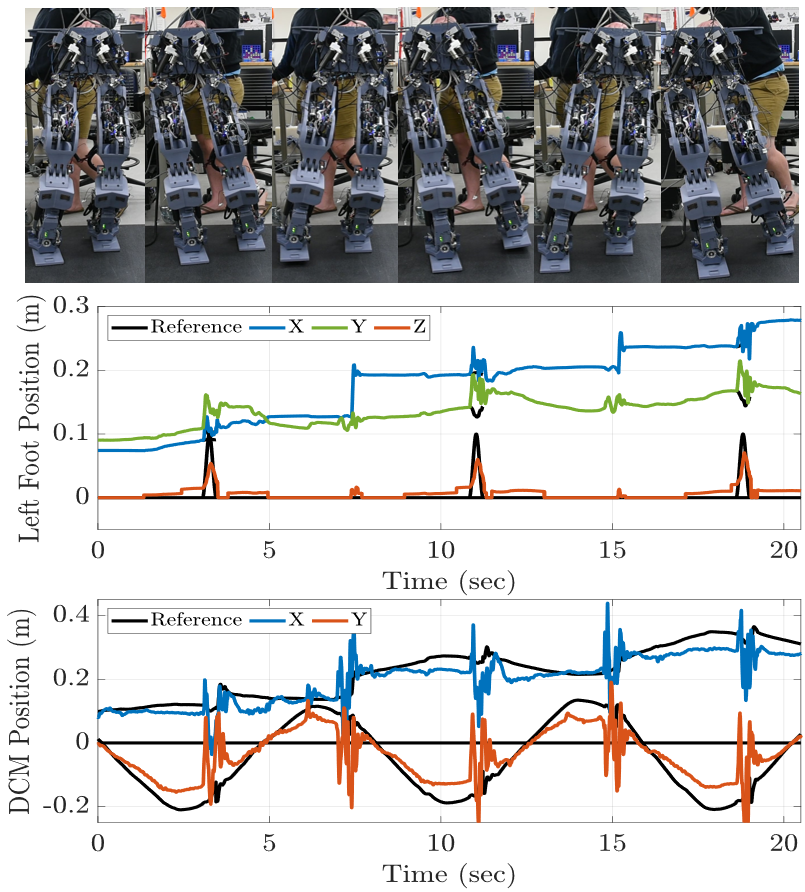}
\caption{Results of the stepping experiment with PANDORA. 5 contiguous footsteps are achieved. DCM and left foot position tracking performance are shown. All reference footstep positions are set to be the same as the actual foot positions when not in swing. }	
\label{fig:Results}
\vspace{-0.5cm}
\end{figure}

Due to these challenges, walking with PANDORA is not yet fully stable. PANDORA can complete several footsteps in a row before falling over, with some stabilization inputs by hand from the operator to damp out the worst vibrations during contact switches.
The results of this stepping experiment are shown in Fig. \ref{fig:Results}. The robot exhibits two major results. First is that the DCM, and thus the COM, tracks the desired positions, but with large instantaneous errors happening during single stance phases. The feet also have difficulties tracking desired positions.
These are the whole-body behavior impacts of the structural elasticity present in each link, which cause the robot to destabilize after the 5 steps shown. Note that the right foot tracking performance is not shown, but it approximately mirrors the behavior of the left foot.

%% file: Sections/Section6_Conclusion.tex
\section{Conclusion}


In this paper we present the overall motivation, design, construction, and performance of the humanoid robot PANDORA. The robot is structurally elastic, meaning that the links are constructed of compliant material. This design has many benefits, such as leveraging rapid and additive manufacturing to design and build the robot and inherently including the actuator elasticity directly into the structural components. However, the structural elasticity presents other challenges, such as difficulty in measuring exact joint torques and positions, and unmodelled elastic motion within the robot causing errors in whole-body momentum and end-effector position estimation. Despite these challenges, we demonstrate that standard whole-body-control approaches can still be implemented to achieve robust balancing and stepping behaviors. Future work will include developing Kalman filter-based approaches to improve estimation of joint motion in the presence of elasticity for improving state estimation and overall control performance. PANDORA will continue to be updated where the full design is available on TREC's public GitLab at \href{https://gitlab.com/trec-lab}{gitlab.com/trec-lab}.